\documentclass[letterpaper, 10 pt, conference]{ieeeconf}  % Comment this line out if you need a4paper

\IEEEoverridecommandlockouts                              % This command is only needed if 
\usepackage{graphicx}
\usepackage{amsmath} % assumes amsmath package installed
\usepackage{amssymb}  % assumes amsmath package installed
\usepackage{blindtext}
\usepackage{xcolor}
\usepackage{url}

\title{\LARGE \bf
Scoring Cycling Environments Perceived Safety using \\Pairwise Image Comparisons
}

\author{Miguel Costa$^{1,2}$ and Manuel Marques$^{2}$ and Felix Wilhelm Siebert$^{3}$ \\ and Carlos Lima Azevedo$^{3}$ and Filipe Moura$^{1}$% <-this % stops a space
\thanks{$^{1}$
        Civil Engineering Research and Innovation for Sustainability, Instituto Superior T\'{e}cnico, Universidade de Lisboa, Av. Rovisco Pais, 1, Lisboa, Portugal
        {\tt\small{fmoura@tecnico.ulisboa.pt}}}%
\thanks{$^{2}$
        Institute for Systems and Robotics, Instituto Superior T\'{e}cnico, Universidade de Lisboa, Av. Rovisco Pais, 1, Lisboa, Portugal {\tt\small{\{mncosta, manuel\}@isr.tecnico.ulisboa.pt}}}%
\thanks{$^{3}$
        Department of Technology, Management and Economics, Technical University of Denmark, Kgs. Lyngby, 2800, Denmark
        {\tt\small{\{felix, climaz\}@dtu.dk}}}%
}

\begin{document}

\maketitle
\thispagestyle{empty}
\pagestyle{empty}

%%%%%%%%%%%%%%%%%%%%%%%%%%%%%%%%%%%%%%%%%%%%%%%%%%%%%%%%%%%%%%%%%%%%%%%%%%%%%%%%
%% ABSTRACT
%%%%%%%%%%%%%%%%%%%%%%%%%%%%%%%%%%%%%%%%%%%%%%%%%%%%%%%%%%%%%%%%%%%%%%%%%%%%%%%%
\begin{abstract}
Today, many cities seek to transition to more sustainable transportation systems. Cycling is critical in this transition for shorter trips, including first-and-last-mile links to transit. Yet, if individuals perceive cycling as unsafe, they will not cycle and choose other transportation modes.
This study presents a novel approach to identifying how the perception of cycling safety can be analyzed and understood and the impact of the built environment and cycling contexts on such perceptions. We base our work on other perception studies and pairwise comparisons, using real-world images to survey respondents. We repeatedly show respondents two road environments and ask them to select the one they perceive as safer for cycling. 
We compare several methods capable of rating cycling environments from pairwise comparisons and classify cycling environments perceived as safe or unsafe. 
Urban planning can use this score to improve interventions' effectiveness and improve cycling promotion campaigns. Furthermore, this approach facilitates the continuous assessment of changing cycling environments, allows for a short-term evaluation of measures, and is efficiently deployed in different locations or contexts.
\end{abstract}

%%%%%%%%%%%%%%%%%%%%%%%%%%%%%%%%%%%%%%%%%%%%%%%%%%%%%%%%%%%%%%%%%%%%%%%%%%%%%%%%
%% INTRODUCTION
%%%%%%%%%%%%%%%%%%%%%%%%%%%%%%%%%%%%%%%%%%%%%%%%%%%%%%%%%%%%%%%%%%%%%%%%%%%%%%%%
\section{INTRODUCTION}
\label{sec:introduction}
% ######################################################
% Cycling
% ######################################################
To promote sustainability, cities worldwide are promoting a transition to public transportation and active transportation. From these, cycling has proven to provide numerous advantages, including benefits to health \cite{gotschi2016cycling}, economy \cite{clifton2013examining}, and reduction of carbon emissions \cite{NEVES2019130}. Despite these benefits, cycling numbers remain predominantly low in some cities. In contrast, barriers to cycling include hilliness, lack of cycling infrastructure, or appropriate bike storage or parking. Yet, the main deterrent to cycling relates to safety concerns \cite{aldred2015investigating, lawson2013perception, felix2019maturing}. If cyclists feel unsafe or are afraid to cycle, they will prefer other means of transportation.

% ######################################################
% Perception of Safety
% ######################################################
Thus, for cities aiming to boost cycling numbers and the effectiveness of such strategies, it is increasingly important to understand what affects individuals' perceptions. Perception of cycling safety research explores how individuals subjectively experience cycling accident risk and what fears and events negatively impact one's perception of being involved in a cycling accident. Current research shows that infrastructure layout, fear of traffic, and distracted cycling are some aspects that influence this perception \cite{heinen2010commuting}. Most research focuses on surveys and in-loco and post-riding interviews to compare factors influencing perceptions \cite{sanders2015perceived}. Even though these approaches are vital to understanding cycling perception of safety, they need to be more scalable over space or time due to their high cost (human resources, time, and money). This prevents any analysis of perceptions over time, and qualitative non-scalable data analysis hampers any comparative study across cities or countries.

\begin{figure}[t]
  \centering
  \includegraphics[width=.49\textwidth, trim={0 2cm 0 0},clip]{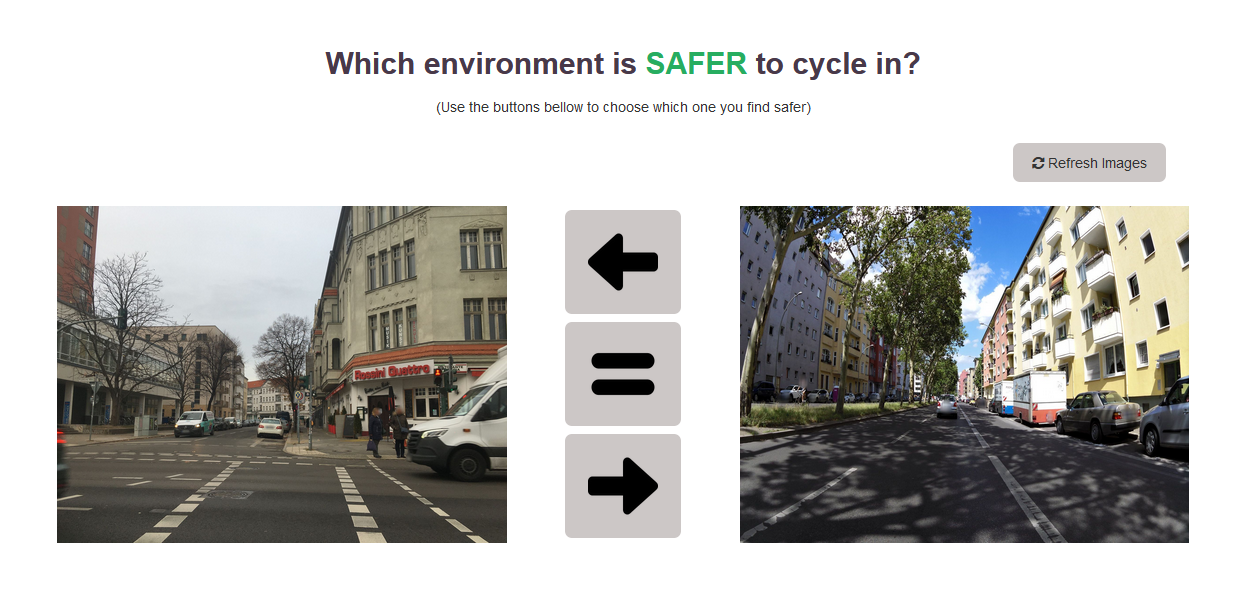}
  \vspace{-18pt}
  \caption{Pairwise image comparison of two cycling environments. Users choose the one (left image, right image, or tie) they find safer for cycling.}
  \label{fig:pairwise_survey}
  \vspace{-18pt}
\end{figure}
% ######################################################
% Pairwise Comparisons
% ######################################################
Studying such perceptions has traditionally been carried out using direct rating methods (users assign a score to each event or situation). This procedure requires a well-defined scale and user training and is particularly difficult to conduct when events or conditions substantially differ from one another \cite{perez2017practical}, which is the case when analyzing real-world environments. In contrast, using pairwise comparisons (users compare two situations and choose one of the two) is often simpler and faster to set up, well-suited for non-expert participants \cite{perez2017practical}, and presents lower measurement error compared to direct ratings \cite{shah2015estimation}. With this in mind, we employ pairwise comparisons to analyze cycling safety perceptions. Moreover, we draw current practice and knowledge from other research areas (e.g., sports outcome prediction and preference learning) about pairwise comparisons and how algorithms can be used to study cycling safety perceptions, something unexplored in cycling safety research. This paves the way to scale safety perception studies and ubiquitously understand how individuals perceive cycling risk.

% ######################################################
% Gap, Objectives & Contributions
% ######################################################
The main contributions of this paper are as follows. First, we draw knowledge from other research areas about pairwise comparisons and apply them to studying cycling safety perceptions. This novel approach
uses a survey showcasing images of two road environments and asking users which one they find safer, if any. % We use respondents' answers to compare different methodologies previously applied to sports prediction and preference learning, showcasing how these can be directly applied to our main goal: understanding cycling perception of safety.
With the respondents' answers, we compare different methodologies, previously applied to sports prediction and preference learning, and show how these can be directly applied to our main goal: understanding cycling perception of safety. Lastly, we draw from these results to objectively classify cycling environments based on urban characteristics and cycling environments.

% ######################################################
% Outline of article
% ######################################################
We divide the article as follows. In the next section, we explore the current literature on pairwise comparisons and how traditional rating methods unravel such data. In Section \ref{sec:survey}, we detail our pairwise comparison survey and present different algorithms to rate cycling environments. Next, in Section \ref{sec:ranking}, we present the methodology, overviewing all pairwise ranking algorithms and environment classification. Section \ref{sec:results} presents the results and highlights what environments are perceived as safer or riskier. Finally, Section \ref{sec:conclusions} concludes the paper and draws possible paths forward.

%%%%%%%%%%%%%%%%%%%%%%%%%%%%%%%%%%%%%%%%%%%%%%%%%%%%%%%%%%%%%%%%%%%%%%%%%%%%%%%%
%% RELATED WORK
%%%%%%%%%%%%%%%%%%%%%%%%%%%%%%%%%%%%%%%%%%%%%%%%%%%%%%%%%%%%%%%%%%%%%%%%%%%%%%%%
\section{RELATED WORK}
\label{sec:related_work}
%%%%%%%%%%%%%%%%%%%%%%%%%%%%%%%%%%%%%%%%%%%%%%%%%%%%%%%%%%%%%%%%%%%%%%%%%%%%%%%%
%% PERCEPTION OF CYCLING SAFETY
%%%%%%%%%%%%%%%%%%%%%%%%%%%%%%%%%%%%%%%%%%%%%%%%%%%%%%%%%%%%%%%%%%%%%%%%%%%%%%%%
\subsection{Subjective cycling safety}
Perceived or subjective safety relates to the feeling of safety of an individual, i.e., how individuals subjectively experience accident risk. Measuring this is vital for municipalities and decision-makers to make informed decisions and adequately provide cyclists with environments they feel safe to cycle in. 
Previous research has found many characteristics that relate to the sense of risk, such as cycling helmets and clothing \cite{aldred2015reframing}, sense of traffic \cite{sanders2015perceived}, urban roads and compliance with road rules \cite{lawson2013perception}, and infrastructure layout \cite{chataway2014safety}.
In effect, urban features can be indexed to measure perceived risk objectively. Indicators and scales, such as the Bicycle Stress Level \cite{sorton1994bicycle}, the Level of Traffic Stress \cite{mekuria2012low} or its updated form \cite{furth2017level}, help planners and researchers to compare contexts and analyze cycling environments.

%%%%%%%%%%%%%%%%%%%%%%%%%%%%%%%%%%%%%%%%%%%%%%%%%%%%%%%%%%%%%%%%%%%%%%%%%%%%%%%%
%% METHODS FOR STUDYING PERCEPTION OF CYCLING SAFETY
%%%%%%%%%%%%%%%%%%%%%%%%%%%%%%%%%%%%%%%%%%%%%%%%%%%%%%%%%%%%%%%%%%%%%%%%%%%%%%%%
The need to acquire such vital data has led researchers to employ qualitative \textit{in situ} or online surveys and interviews to understand what urban features may trigger or negatively arouse individuals \cite{sanders2015perceived, aldred2015reframing}. Naturalistic and semi-naturalistic approaches are often used. These approaches focus on more quantitative methods to capture human responses to risky environments, such as using physiological data using wearable sensors \cite{zeile2016urban}, showcasing cycling videos \cite{Parkin2007}, use of virtual reality \cite{von2018risk}, or eye tracking devices \cite{schmidt2018risk}. 
Yet, these approaches are often not scalable as they are time-consuming and resource-intensive, require precise preparation and monitoring of special devices, or may require individual training.

Recently, some methods have been proposed to counter this. For example, \cite{von2022safe} used a Likert-scale-based survey using 1900 images of cycling environments to generalize recommendations regarding best practices regarding subjectively safe cycling lanes. 
\cite{ito2021assessing} have used computer vision to index bikeability utilizing several automatically extracted features from street-view images (SVI) to compare Tokyo and Singapore. 
Likewise, although not applied to cycling safety, machine learning, and other data processing methodologies have explored how individuals perceive different environments, enabling faster, easier, and automatic evaluations for different perceptions \cite{Naik2014, dubey2016deep, ramirez2021measuring}.

%%%%%%%%%%%%%%%%%%%%%%%%%%%%%%%%%%%%%%%%%%%%%%%%%%%%%%%%%%%%%%%%%%%%%%%%%%%%%%%%
%% PAIRWISE COMPARISONS
%%%%%%%%%%%%%%%%%%%%%%%%%%%%%%%%%%%%%%%%%%%%%%%%%%%%%%%%%%%%%%%%%%%%%%%%%%%%%%%%
\subsection{Pairwise comparisons}
Pairwise comparison models aim to predict the outcome of comparing two items, i.e., when items $A$ and $B$ are compared, would a user prefer item $A$, item $B$, or would they be perceived equally (tie)? These models were first proposed in psychophysics and marketing research and have typically followed the seminal works of Thurstone \cite{thurstone1927law} and Bradley–Terry \cite{bradley1952rank}. In the past decades, paired comparison models have been explored and applied to many domains, including sports skill ranking and game prediction \cite{maystre2019pairwise, chau2023spectral}, image quality analysis \cite{xu2016pairwise}, and city perceptions \cite{Naik2014, costa2019citysafe}.

%%%%%%%%%%%%%%%%%%%%%%%%%%%%%%%%%%%%%%%%%%%%%%%%%%%%%%%%%%%%%%%%%%%%%%%%%%%%%%%%
%% RATING METHODS FROM PAIRWISE COMPARISONS
%%%%%%%%%%%%%%%%%%%%%%%%%%%%%%%%%%%%%%%%%%%%%%%%%%%%%%%%%%%%%%%%%%%%%%%%%%%%%%%%
Typical models assume that there is a latent score $s_i$ for each item $i$ and the outcome probability on a comparison between items $i$ and $j$ is a function of the difference between their scores, e.g., $\theta(s_i-s_j)$. Models' usual underlying goal is to estimate the latent scores $s_i$ from the data to obtain an interpretable and comparable score for each item. If $s_i > s_j$, a user would have a greater probability of picking item $i$. The function $\theta$ can have many forms but usually follows a Gaussian or logistic distribution initially proposed by Thurstone \cite{thurstone1927law} and Bradley–Terry \cite{bradley1952rank}, respectively.

Several methodologies have been proposed to extend comparison models, including iterative algorithms, Bayesian-based models, and covariate-based or covariate-free models. Covariate-based models often allow for new items to be added to the comparison set seamlessly without any prior comparison involving new items. Yet, these methods require having said covariates and do not rely entirely on the outputs of paired comparisons. For this work, we focus on covariate-free models requiring only results from pairwise comparisons.
For iterative algorithms, probably the most well-known case is the Elo rating \cite{elo1978rating}, which has been used to rank chess players by FIDE\footnote{https://ratings.fide.com/calc.phtml?page=change}, by FIFA to rank women's national football teams\footnote{https://www.fifa.com/fifa-world-ranking/procedure-women}, or by FiveThirtyEight to rank NFL teams\footnote{https://fivethirtyeight.com/methodology/how-our-nfl-predictions-work/}. Elo uses a simple online stochastic update rule based on an item's scores and the expected outcome of one item winning over the other. Despite its simplicity, Elo has remained one of the most used procedures since it is tractable and can easily adjust to diverse situations and scenarios. For Bayesian models, both Glicko \cite{glickman1999parameter} and TrueSkill \cite{herbrich2006trueskill} have been put forward as probabilistic methods that measure not only the latent scores $s_i$ but also the uncertainty associated with each score, which is often valuable.

More recently, other approaches have been suggested using alternative methodologies. These include spectral ranking that (usually) involves computing the pairwise comparison matrix leading eigenvalues and eigenvectors \cite{chau2023spectral}, convex problem formulation that usually penalizes wrongly or contradictory answers \cite{xu2016pairwise, costa2019citysafe}, or Gaussian processes to model different data dynamics \cite{maystre2019pairwise, chu2005preference}. 

In this work, we study paired comparison models to analyze cycling perception of safety. To the best of our knowledge, this has not been explored before and can potentially help researchers analyze the impact of the cycling environments on individuals' perceptions, enabling faster and continuous evaluations of such effects.

%%%%%%%%%%%%%%%%%%%%%%%%%%%%%%%%%%%%%%%%%%%%%%%%%%%%%%%%%%%%%%%%%%%%%%%%%%%%%%%%
%% PAIRWISE IMAGE COMPARISON
%%%%%%%%%%%%%%%%%%%%%%%%%%%%%%%%%%%%%%%%%%%%%%%%%%%%%%%%%%%%%%%%%%%%%%%%%%%%%%%%
\section{PAIRWISE IMAGE COMPARISON SURVEY}
\label{sec:survey}
Aimed at capturing individuals' perceptions of risk, we create a two-part survey. The first part aimed to collect information regarding the user's cycling profile and followed a slightly modified version survey on cyclists' typologies \cite{dill2013four}. Going forward, we focus solely on the second part, which employs pairwise image comparisons of cycling environments. Instituto Superior Técnico's Ethics Committee evaluated the survey, which we then deployed online. The survey took about 10-15 minutes to complete. 

\begin{figure}[t]
  \centering
  \includegraphics[width=.23\textwidth, trim={0 220pt 0 0}, clip]{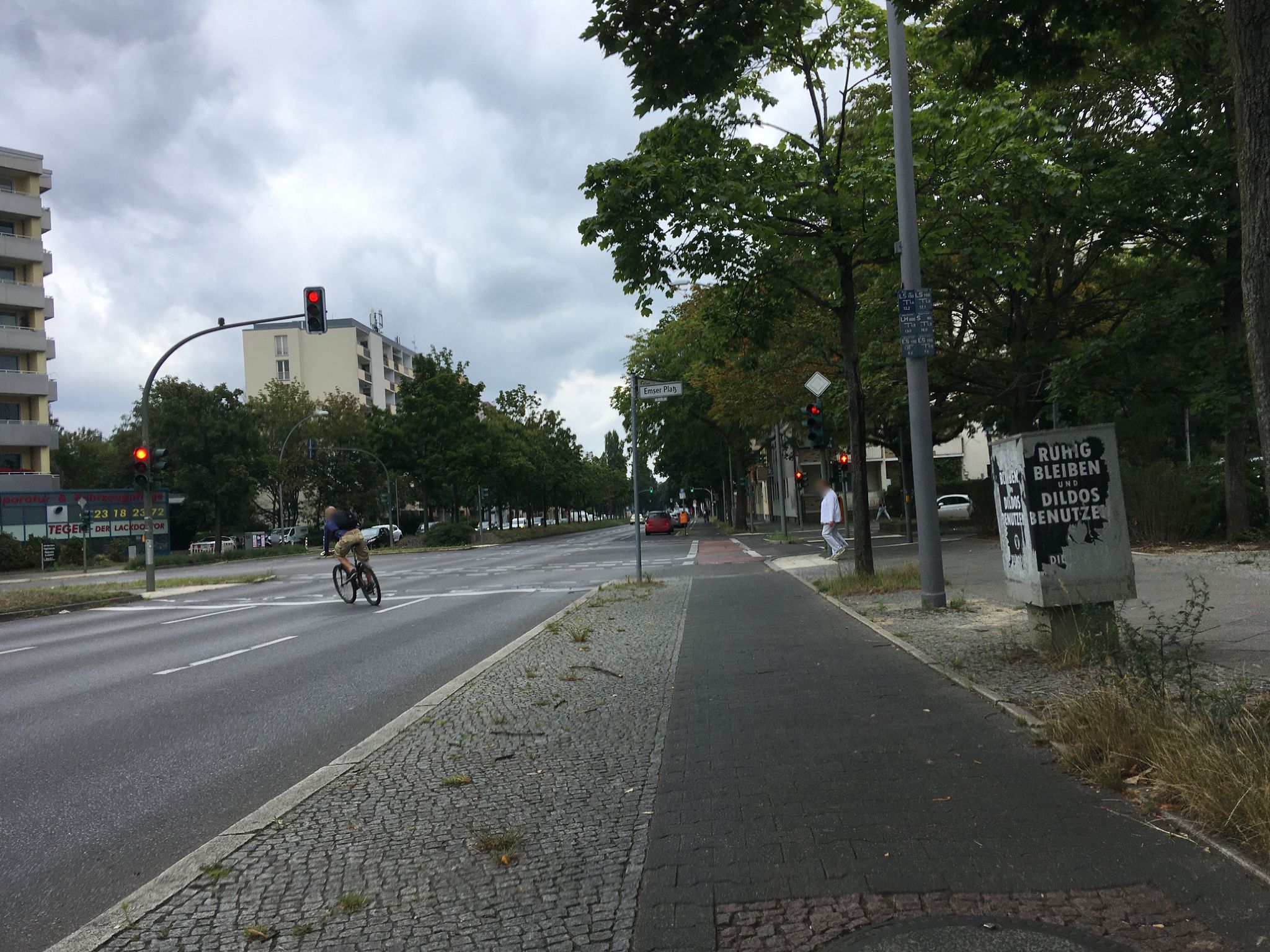}%
  \includegraphics[width=.23\textwidth, trim={0 220pt 0 0}, clip]{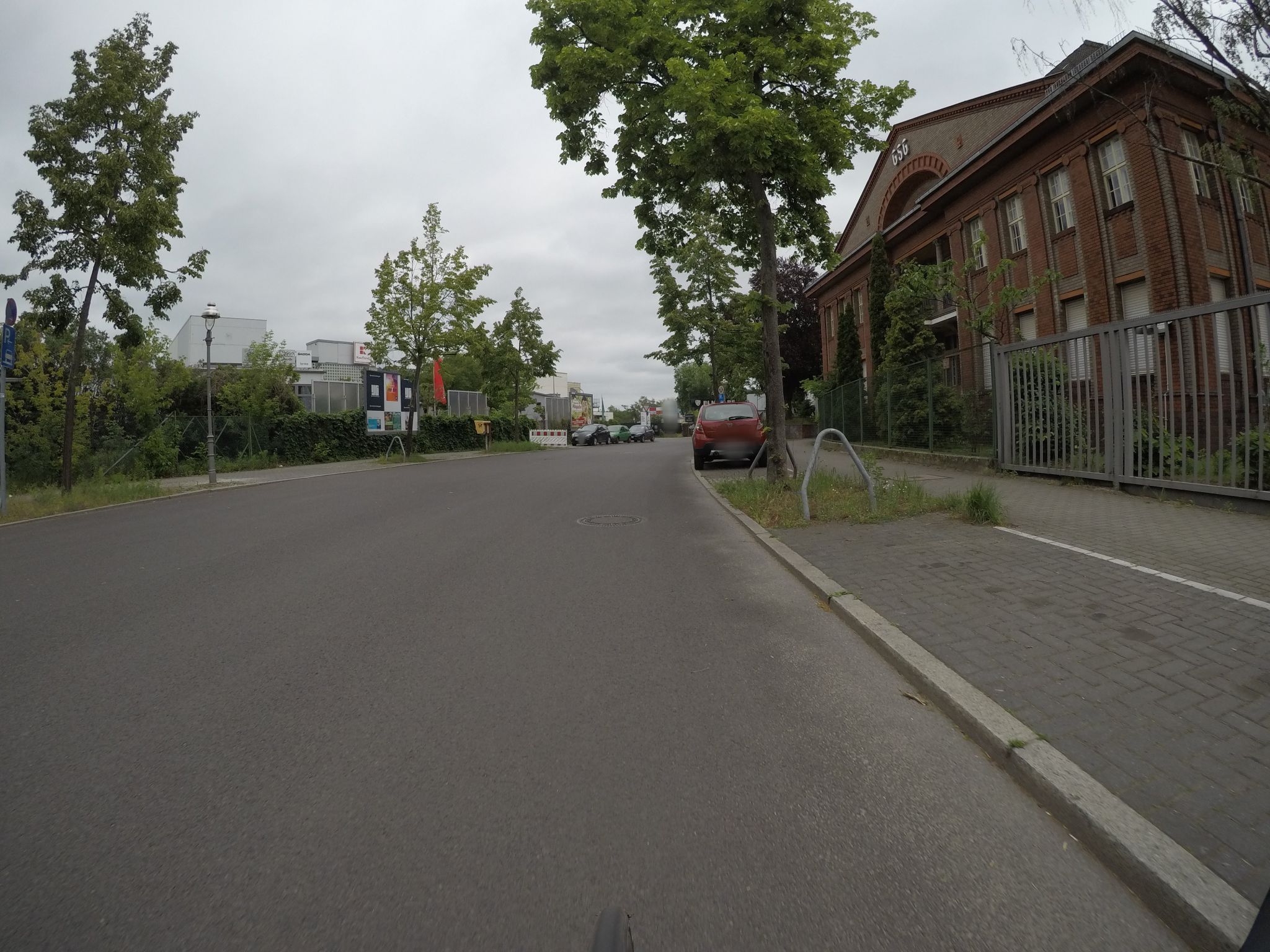}\\
  \includegraphics[width=.23\textwidth, trim={0 220pt 0 0}, clip]{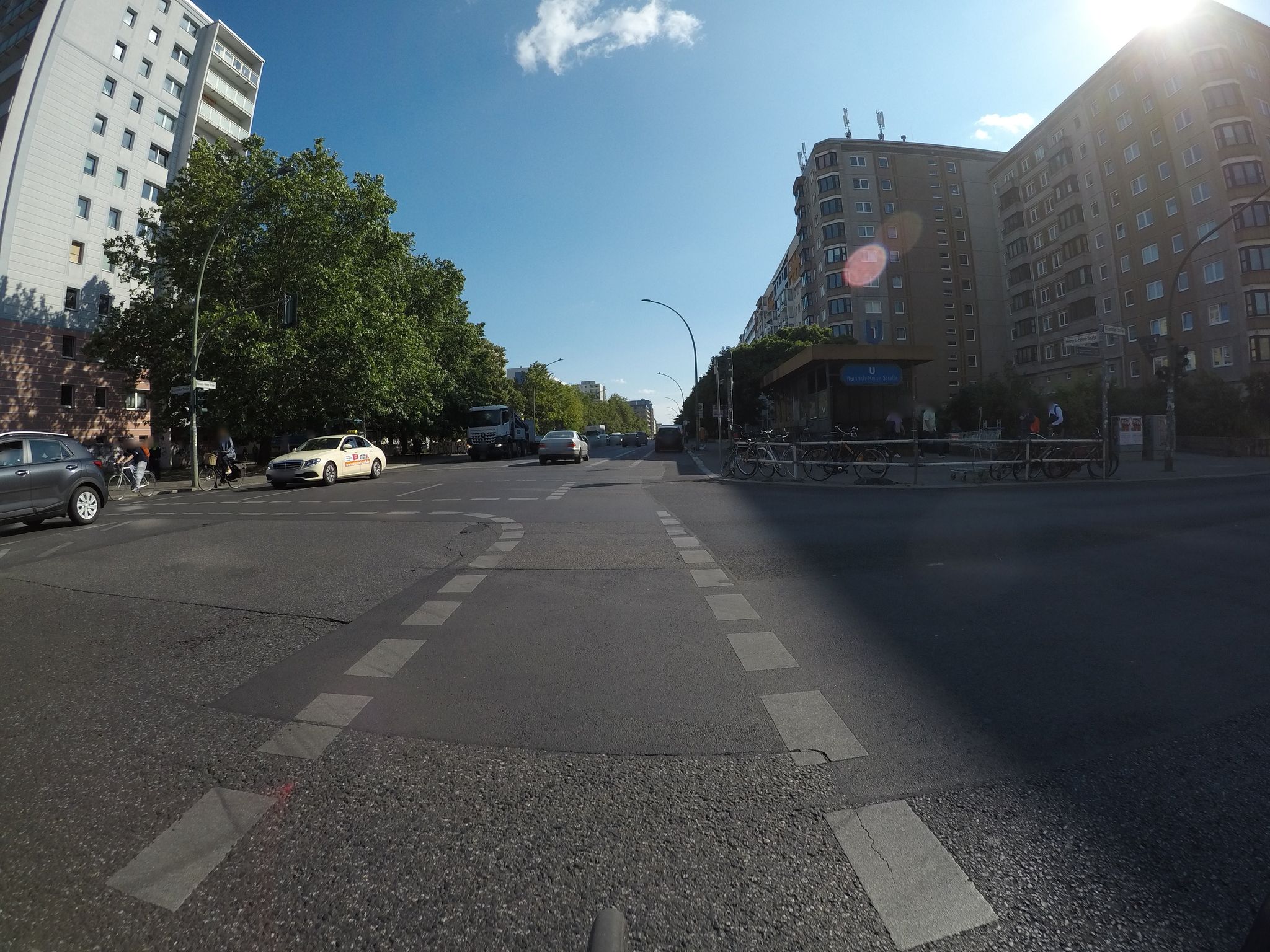}%
  \includegraphics[width=.23\textwidth, trim={0 220pt 0 0}, clip]{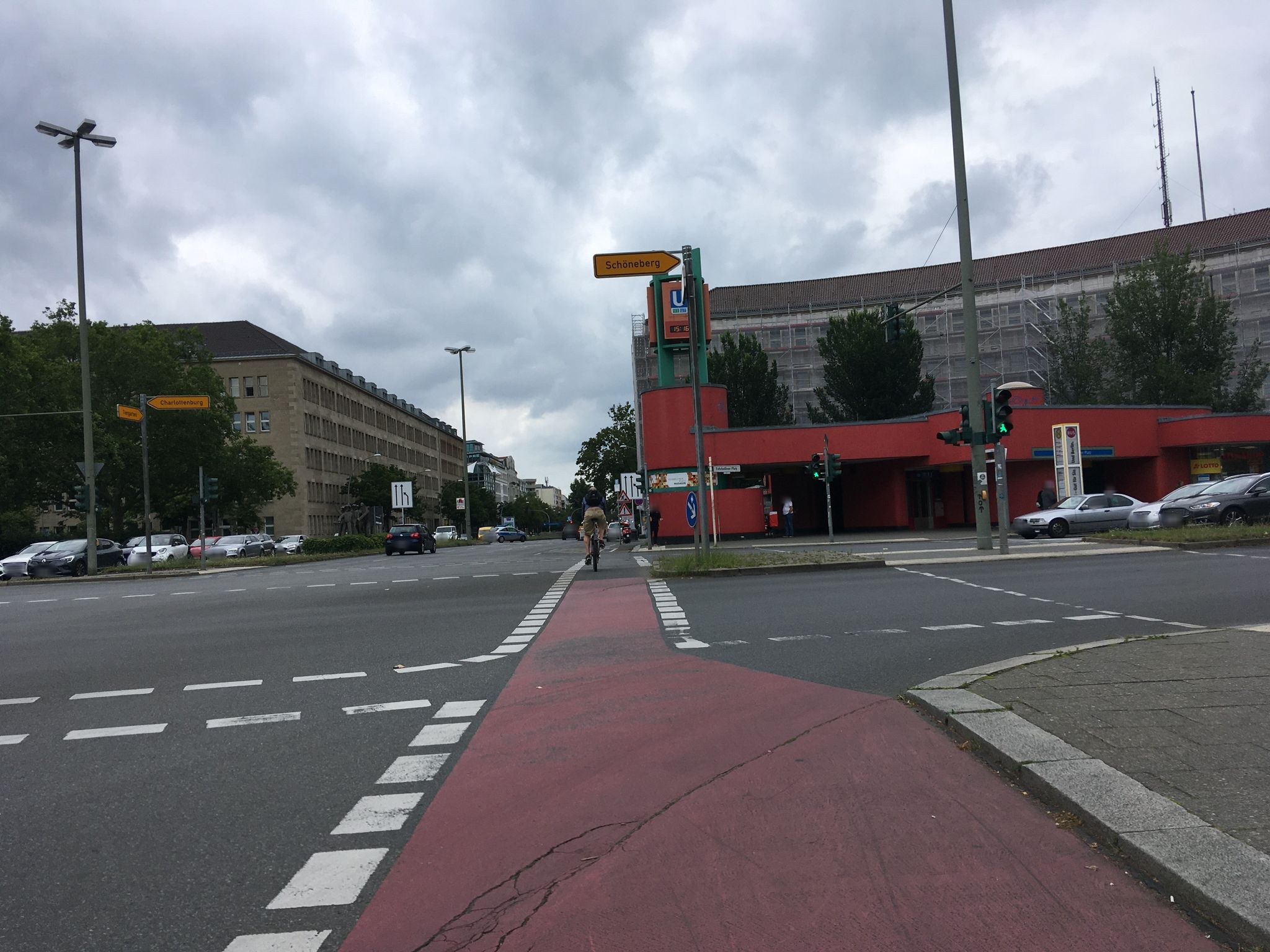}\\
  \vspace{-10pt}
  \caption{Example of four cycling environments used in our image pairwise comparison survey.}
  \label{fig:images_example}
  \vspace{-12pt}
\end{figure}

%%%%%%%%%%%%%%%%%%%%%%%%%%%%%%%%%%%%%%%%%%%%%%%%%%%%%%%%%%%%%%%%%%%%%%%%%%%%%%%%
%% Pairwise image comparisons
%%%%%%%%%%%%%%%%%%%%%%%%%%%%%%%%%%%%%%%%%%%%%%%%%%%%%%%%%%%%%%%%%%%%%%%%%%%%%%%%
%\subsection{Pairwise image comparisons}
We repeatedly present respondents with two road environment pictures and ask them to select the one they perceive as safer for cycling (Figure \ref{fig:pairwise_survey}). We randomly sampled street-view images of road environments from Mapillary (https://www.mapillary.com/) from Berlin, Germany. 
The selected array of pictures captures a wide range of urban environments, including different infrastructure layouts, dedicated cycle lanes, urban characteristics, street furniture, vegetation, and varying degrees of other road users and pedestrians. 
We collected a set of 4481 total images across Berlin. Figure \ref{fig:images_example} shows some cycling environment images. To show pairs of pictures to respondents, we preprocess each image to extract key attributes about the depicted environment. We employ a partial factorial design, randomly selecting two photos with the same level of features, while others are free to vary, e.g., both images have the same level of vegetation, or both include a cycle lane. We ask respondents to complete 65 paired comparisons, but they can stop at an earlier number.

We collect responses from 251 users, averaging 3.25 comparisons per image and 29 comparisons per respondent. Of the respondents, 123 identified as males, 71 as females, with the remaining preferring not to disclose their gender. Agewise, 86 were between the ages of 18-30, 64 between 31-40, 30 between 41-50, 16 mentioned they were older than 51, and the remaining did not specify any age. Overall, individuals could be classified according to Geller's cycling profiles \cite{geller2006four} as No Way, No How (5.1\%); Interested, but Concerned (51.5\%); Enthused \& Confident (38.3\%); and Strong \& Fearless (4.1\%).

%%%%%%%%%%%%%%%%%%%%%%%%%%%%%%%%%%%%%%%%%%%%%%%%%%%%%%%%%%%%%%%%%%%%%%%%%%%%%%%%
%% METHODOLOGY
%%%%%%%%%%%%%%%%%%%%%%%%%%%%%%%%%%%%%%%%%%%%%%%%%%%%%%%%%%%%%%%%%%%%%%%%%%%%%%%%
\section{PERCEPTION OF SAFETY SCORES}
\label{sec:ranking}
%%%%%%%%%%%%%%%%%%%%%%%%%%%%%%%%%%%%%%%%%%%%%%%%%%%%%%%%%%%%%%%%%%%%%%%%%%%%%%%%
%% PERCEPTION OF SAFETY SCORES
%%%%%%%%%%%%%%%%%%%%%%%%%%%%%%%%%%%%%%%%%%%%%%%%%%%%%%%%%%%%%%%%%%%%%%%%%%%%%%%%
\subsection{Computing scores}
After preparing and deploying the pairwise image comparison survey, we explore and compare covariate-free methodologies to compute subjective safety scores. This score allows non-experts and decision-makers to understand and compare cycling environments easily. We now provide an overview of each method.

%%%%%%%%%%%%%%%%%%%%%%%%%%%%%%%%%%%%%%%%%%%%%%%%%%%%%%%%%%%%%%%%%%%%%%%%%%%%%%%%
%% ELO
%%%%%%%%%%%%%%%%%%%%%%%%%%%%%%%%%%%%%%%%%%%%%%%%%%%%%%%%%%%%%%%%%%%%%%%%%%%%%%%%
\noindent \textbf{Elo \cite{elo1978rating}} We start by setting an initial score $s_0$ for each image. Next, after each comparison, we compute the expected result for item $A$ between items $A$ and $B$:
\begin{equation}
E_A = \frac{1}{1+10^{(s_B-s_A)/\delta}},
\label{eq:elo_expected}
\end{equation}
with $\delta$ modulating the scores difference. The update score for item $A$, $s_A'$, can then be updated using the following:
\begin{equation}
s_A' = s_A + k (\gamma - E_A),
\label{eq:elo_update}
\end{equation}
with $k$ modulating the impact of the outcome on the new score and $\gamma$ being 1 for the winning item and 0 for the losing one, or 0.5 for ties for both items.

%%%%%%%%%%%%%%%%%%%%%%%%%%%%%%%%%%%%%%%%%%%%%%%%%%%%%%%%%%%%%%%%%%%%%%%%%%%%%%%%
%% TRUESKILL
%%%%%%%%%%%%%%%%%%%%%%%%%%%%%%%%%%%%%%%%%%%%%%%%%%%%%%%%%%%%%%%%%%%%%%%%%%%%%%%%
\noindent \textbf{TrueSkill (TS) \cite{herbrich2006trueskill}} This Bayesian framework assumes that each image's score is modeled by a $\mathcal{N}(\mu,\,\sigma^{2})$ random variable, which is updated after each comparison. Update rules follow that, for image $A$ winning over image $B$:
\begin{equation}
\begin{aligned}
\mu_A' = \mu_A + \frac{\sigma_A^2}{c} \cdot f(\frac{\mu_A-\mu_B}{c}, \frac{\varepsilon}{c})\\ 
\mu_B' = \mu_B + \frac{\sigma_B^2}{c} \cdot f(\frac{\mu_A-\mu_B}{c}, \frac{\varepsilon}{c})\\
\sigma_A^{2'} = \sigma_A^2 (1-\frac{\sigma_A^2}{c} \cdot g(\frac{\mu_A-\mu_B}{c}, \frac{\varepsilon}{c})) \\
\sigma_B^{2'} = \sigma_B^2 (1-\frac{\sigma_B^2}{c} \cdot g(\frac{\mu_A-\mu_B}{c}, \frac{\varepsilon}{c})) \\
c^2 = 2\beta^2 + \sigma_A^2 + \sigma_B^2
\end{aligned},
\label{eq:trueskill_update}
\end{equation}
with $\beta$ being a per-game variance parameter, $\varepsilon$ an empirical probability of a comparison resulting in a tie, functions $f(\theta) = \mathcal{N}(\theta)/\Phi(\theta)$ and $g(\theta) = f(\theta)\cdot(f(\theta)+\theta)$ defined as the Gaussian density function $\mathcal{N}(\theta)$ and Gaussian cumulative density function $\Phi(\theta)$.

%%%%%%%%%%%%%%%%%%%%%%%%%%%%%%%%%%%%%%%%%%%%%%%%%%%%%%%%%%%%%%%%%%%%%%%%%%%%%%%%
%% CONVEX OPTIMIZATION
%%%%%%%%%%%%%%%%%%%%%%%%%%%%%%%%%%%%%%%%%%%%%%%%%%%%%%%%%%%%%%%%%%%%%%%%%%%%%%%%
\noindent \textbf{Convex Optimization (CO)} To model paired comparisons, we solve a convex optimization program following \cite{costa2019citysafe}:
\begin{equation}
\begin{aligned}
& \underset{s, t}{\text{minimize}}
& & 1^Tt +\lambda_{ties}1^T|B^Ts|\\
& \text{subject to}
& & 1^Ts=0\\
& & & \epsilon-b_n^Ts \le t_n\\
& & & 0 \le t_n, n=1, ..., N
\end{aligned},
\label{eq:cvx_opt}
\end{equation}
with $s\in\mathbb{R}^M$ being the score vector for $M$ images, $N$ the total number of comparisons, $b_n$ a vector containing information for comparing pairs ($b_n$ is a vector of zeros, with 1 in the $m$-th position of the winning image, and -1 in the $m$-th position of the losing one), and $\epsilon$ an error margin to tolerate offending comparisons. This cost function penalizes scores that violate the error margin greater than $\epsilon$. The optimal scores $s$ will be the one that violates the least paired comparisons and, if so, the ones where image scores are closer.

%%%%%%%%%%%%%%%%%%%%%%%%%%%%%%%%%%%%%%%%%%%%%%%%%%%%%%%%%%%%%%%%%%%%%%%%%%%%%%%%
%% GAUSSIAN PROCESS
%%%%%%%%%%%%%%%%%%%%%%%%%%%%%%%%%%%%%%%%%%%%%%%%%%%%%%%%%%%%%%%%%%%%%%%%%%%%%%%%
\noindent \textbf{Gaussian Process (GP)} We perform approximate Bayesian inference over pairwise comparisons following \cite{maystre2019pairwise}. Here, scores are approximated by a Gaussian Process ($s(n) \sim \mathcal{GP}(0, k(n, n'))$) defined by the joint distribution of N pairwise comparisons of scores $s$, with $s \sim \mathcal{N}(0, K)$, with $K$ being the covariance matrix $K=[k(n_i, n_j)]$, defined by a covariance function that models the dynamics of scores over comparisons. We chose a logit observation model and defined the likelihood accordingly. For further detail on the approximate posterior probabilities and inference through Expectation-Propagation, we refer the reader to \cite{maystre2019pairwise}.

%%%%%%%%%%%%%%%%%%%%%%%%%%%%%%%%%%%%%%%%%%%%%%%%%%%%%%%%%%%%%%%%%%%%%%%%%%%%%%%%
%% LUCE SPECTRAL RANKING
%%%%%%%%%%%%%%%%%%%%%%%%%%%%%%%%%%%%%%%%%%%%%%%%%%%%%%%%%%%%%%%%%%%%%%%%%%%%%%%%
\noindent \textbf{Luce Spectral Ranking (LSR) \cite{maystre2015fast}} By constructing pairwise comparisons as a graph, where edges represent comparisons and their results, this algorithm works as a scoring function of such graph representation. The graph's structure defines probabilities as the stationary probability of a natural random walk over nodes (images) or a stationary distribution of a Markov chain. Essentially, this measures the likelihood of moving from item $A$ to item $B$, which depends on how many comparisons item $A$ won versus item $B$. As such, it captures an item's preference globally over all other items.

%%%%%%%%%%%%%%%%%%%%%%%%%%%%%%%%%%%%%%%%%%%%%%%%%%%%%%%%%%%%%%%%%%%%%%%%%%%%%%%%
%% PREDICTING SUBJECTIVE SAFETY SCORES
%%%%%%%%%%%%%%%%%%%%%%%%%%%%%%%%%%%%%%%%%%%%%%%%%%%%%%%%%%%%%%%%%%%%%%%%%%%%%%%%
\subsection{Predicting environments as perceived safe or unsafe}
After scoring each cycling environment, we aim to predict if environments are perceived as safe or unsafe based on image characteristics. This classification can help urban planners and designers to understand what urban features impact individuals' cycling perception of accident risk. 

As such, we perform binary classification to classify environments as perceived \textit{safe} or \textit{unsafe}. To get a representation of the image, we run images through the widely popular deep neural network InceptionV3 \cite{szegedy2016rethinking} pre-trained on ImageNet, from which we remove the final softmax classification layer. Other architectures were tested, with InceptionV3 providing the best results for this task. From this, we extract a latent representation of the urban environment for each image to be used as the predictor in our classification problem.

Next, we label environments as \textit{safe} or \textit{unsafe} by setting a threshold on the predicted rating using one of the algorithms from Section \ref{sec:ranking}. We set $s_H$ and $s_L$, where images with a score above $s_H$ are perceived as \textit{safe}, and below $s_L$ are perceived as unsafe. These thresholds are defined as $s_H= \bar{s} + \alpha \sigma_s$ and $s_L= \bar{s} - \alpha \sigma_s$, with $\bar{s}$ and $\sigma_s$ being the average and standard deviation of the scores on the test set, respectively, and $\alpha$ a varying parameter set to control how distant perceived safer environments are from unsafe ones. Particularly, if $\alpha=0$, then $s_H=s_L=\bar{s}$, meaning that their environments are either perceived as safe or unsafe.
Finally, we use eXtreme Gradient Boosting Tree (XGBoost) \cite{chen2016xgboost} to perform binary classification due to being a powerful approach to binary classification.

%%%%%%%%%%%%%%%%%%%%%%%%%%%%%%%%%%%%%%%%%%%%%%%%%%%%%%%%%%%%%%%%%%%%%%%%%%%%%%%%
%% 
%%%%%%%%%%%%%%%%%%%%%%%%%%%%%%%%%%%%%%%%%%%%%%%%%%%%%%%%%%%%%%%%%%%%%%%%%%%%%%%%
\section{RESULTS}
\label{sec:results}
This section details the results of modeling pairwise comparisons using the methodologies above. We begin by presenting implementation details. Next, we present the results for each paired comparison model and the information about predicting environment perception scores based on environment characteristics.

%%%%%%%%%%%%%%%%%%%%%%%%%%%%%%%%%%%%%%%%%%%%%%%%%%%%%%%%%%%%%%%%%%%%%%%%%%%%%%%%
%% IMPLEMENTATION DETAILS
%%%%%%%%%%%%%%%%%%%%%%%%%%%%%%%%%%%%%%%%%%%%%%%%%%%%%%%%%%%%%%%%%%%%%%%%%%%%%%%%
We begin by splitting the available pairwise comparisons into train and test sets (85-15\% split). We run a grid search for each model over tunable hyperparameters and present results for the best model. Table \ref{tab:hyperparameters} shows the best hyperparameters. To evaluate each method, we compute the negative average logarithmic loss:\vspace{-3pt}
\begin{equation}
\textsf{log loss} = -\frac{1}{N}\sum^{N}_{n=1} log(p(y^*)),
\label{eq:metrics_logloss}
\end{equation}
for pairwise comparison output $y^*$, and average accuracy. We note that a 
random predictor's accuracy would be 50\%. Log loss provides a good gauge of model calibration, heavily penalizing models for outcomes it considers improbable. We report evaluation metrics on the test set, averaged over five different seeds.
All models were implemented in Python and are publicly available online\footnote{\url{https://github.com/mncosta/scoring_pairwise}}.

\begin{table}[]
\centering
\caption{Hyperparameters used in the paired comparison and classification models.}
\label{tab:hyperparameters}
\vspace{-6pt}
\begin{tabular}{lc}
Model                 & Hyperparameters \\
\hline
Elo                   & $\gamma=400, k=32, s_0=1500$                   \\
TrueSkill             & $s_0=25, \sigma_0=8.33, \beta=4.17, \varepsilon=0.1$                   \\
Convex Optimization   & $\epsilon=0.1$                   \\
Gaussian Process      & $\mathsf{Tie\:margin}=2$                   \\
\hline  \hline    
XGBoost      &  $\mathsf{Max\:depth}=2$, $\mathsf{N\:Estimators}=105$, \\
  & $\mathsf{Learning\:Rate}=0.01$, $\mathsf{Subsample}=0.5$, \\
  & $\mathsf{Feature\:sample\:by\:tree}=0.5$                 \\
\hline
\end{tabular}
%\vspace{-16pt}
\end{table}

\begin{figure*}[t]
  \centering
  \includegraphics[width=.98\textwidth,]{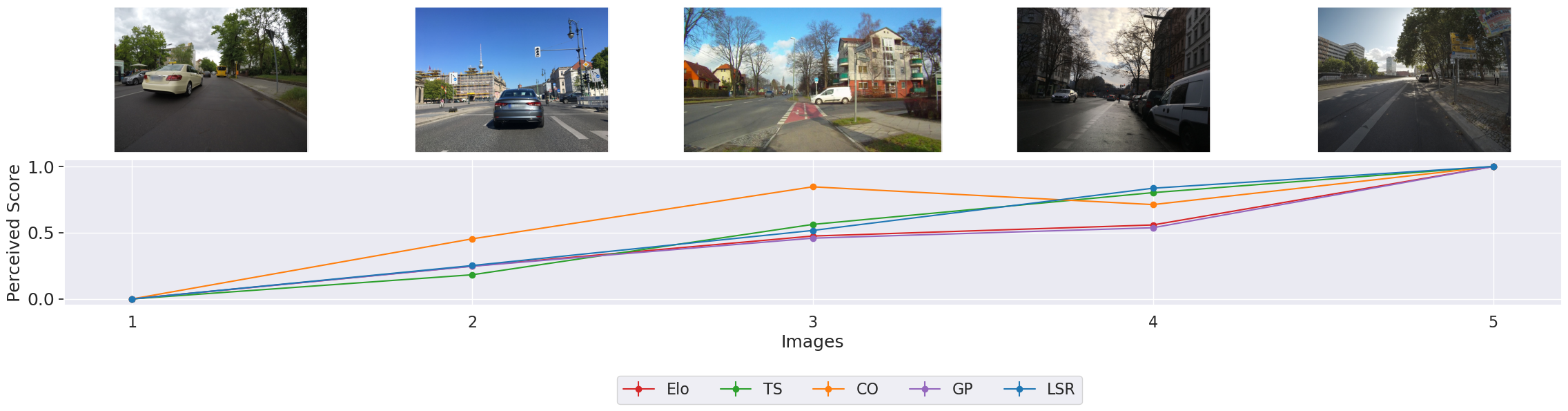}\\
  \vspace{-10pt}
  \caption{Normalized perceived cycling safety scores for all paired comparison models.}
  \label{fig:sequence_scores}
  \vspace{-12pt}
\end{figure*}

%%%%%%%%%%%%%%%%%%%%%%%%%%%%%%%%%%%%%%%%%%%%%%%%%%%%%%%%%%%%%%%%%%%%%%%%%%%%%%%%
%% CYCLING ENVIRONMENT RATING
%%%%%%%%%%%%%%%%%%%%%%%%%%%%%%%%%%%%%%%%%%%%%%%%%%%%%%%%%%%%%%%%%%%%%%%%%%%%%%%%
\subsection{Cycling environment rating}

\begin{table}[]
\centering
\caption{Evaluation metrics for each pairwise model.}
\label{tab:models_results}
\vspace{-6pt}
\begin{tabular}{lrr}
Model    & Log loss $\downarrow$ & Accuracy $\uparrow$\\
\hline
Elo                   &  0.658     &  0.658      \\
TrueSkill             &  0.630     &  0.667      \\
Convex Optimization   &  0.774     &  0.599      \\
Gaussian Process      &  0.839     &  \textbf{0.671}     \\
Luce Spectral Ranking &  \textbf{0.623}     &  0.654 	 \\
\hline
\end{tabular}
\vspace{-12pt}
\end{table}

Table \ref{tab:models_results} shows each model's log loss and accuracy. LSR reveals the lowest log loss but with values close to the TS's. In turn, GP showcases the highest accuracy but with a log loss much higher than that of LSR, meaning that, while it is more accurate, its probability of choosing the winning environment is usually much lower than that of TS or LSR.
We depict the normalized predicted perceived safety scores in Figure~\ref{fig:sequence_scores} for all models, with higher values representing environments perceived as safer. All methods show similar perceived safety score trends, showcasing the lowest scores for the same environments and similar tendencies for the perceived safer ones. We highlight some characteristics by visually inspecting each environment and its predicted score. First, images with non-parked cars (Images 1 and 2) show the lowest score, indicating that the presence of these vehicles decreases the perception of safety. Image 5 has the highest perceived safety score showing a cycle lane and no cars in sight. Images 3 and 4 show average to high scores. While Image 3 shows a cycling lane, it also shows an intersection with other vehicles crossing it. In turn, Image 4 was not taken in an intersection, which was perceived as slightly safer. Additionally, lighting conditions and slight lens distortion play no role in individuals' perception, and only semantic and urban characteristics seem to influence perceptions score.

%%%%%%%%%%%%%%%%%%%%%%%%%%%%%%%%%%%%%%%%%%%%%%%%%%%%%%%%%%%%%%%%%%%%%%%%%%%%%%%%
%% CYCLING ENVIRONMENT RATING
%%%%%%%%%%%%%%%%%%%%%%%%%%%%%%%%%%%%%%%%%%%%%%%%%%%%%%%%%%%%%%%%%%%%%%%%%%%%%%%%
\subsection{Binary classification}

Lastly, we aim to understand if cycling environments can be predicted to be perceived as either safe or unsafe directly from image features. We use XGBoost to perform binary classification on cycling environments, tuning hyperparameters using grid search over a 5-fold cross-validation procedure. Optimal hyperparameters are shown in Table~\ref{tab:hyperparameters}. Given its relatively high accuracy and low log loss, we perform classification using TS scores. To decrease the impact of pictures with few comparisons, we conduct classification only on images whose certainty has reduced past $1/6$ of the initial $\sigma$ value. Images with scores within $[s_L, s_H]$ are considered neutral and removed from this analysis.

\begin{figure}[!htb]
  \centering
  \includegraphics[width=.49\textwidth]{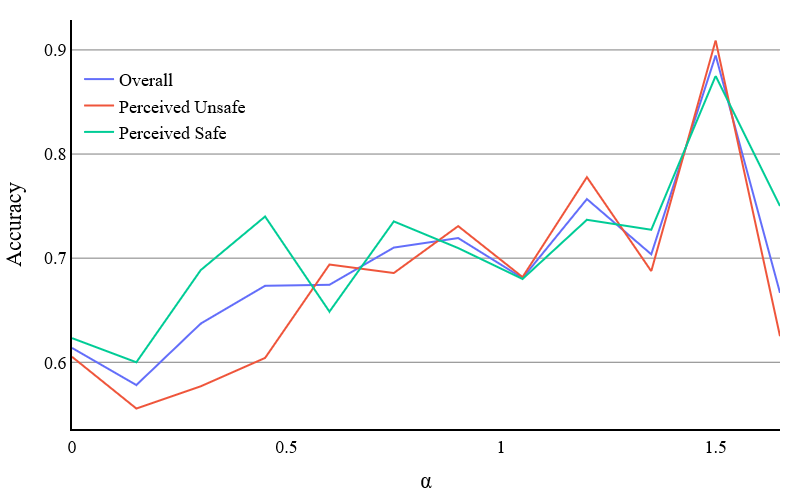}
  \vspace{-18pt}
  \caption{Classification accuracy of \textit{perceived safe} and \textit{perceived unsafe} cycling environments as a function of $\alpha$.}
  \label{fig:binary_classification_alpha}
  \vspace{-18pt}
\end{figure}

Classification accuracy is shown in Figure~\ref{fig:binary_classification_alpha}. When $\alpha=0$, the model has 61.4\% accuracy, reaching an accuracy of 89.5\% when $\alpha=1.5$. While increasing the value of $\alpha$ limits the grouping of environments being perceived as safe or unsafe, it also increases the distinction between the two classes, thus increasing the model's accuracy. For urban planners who seek to massively understand how their cities impact cyclists' perception of risk, this process can be widely adapted to analyze a city's urban form and infrastructure.

%%%%%%%%%%%%%%%%%%%%%%%%%%%%%%%%%%%%%%%%%%%%%%%%%%%%%%%%%%%%%%%%%%%%%%%%%%%%%%%%
%% CONCLUSIONS
%%%%%%%%%%%%%%%%%%%%%%%%%%%%%%%%%%%%%%%%%%%%%%%%%%%%%%%%%%%%%%%%%%%%%%%%%%%%%%%%
\section{CONCLUSIONS}
\label{sec:conclusions}
In this work, we have explored a novel methodology to analyze the perception of cycling safety using pairwise image comparisons. We explore and compare different popular covariate-free paired models to rate cycling environments according to individuals' perceptions, achieving good accuracies for the total number of comparisons. In addition, we explore how binary classification can be used to classify environments as being perceived as safe or unsafe directly from image characteristics. The results show this methodology's potential for widely comparing cycling environments and understanding how these environments impact individuals' perceptions of risk. Moreover, even with few comparisons, the information extracted is very relevant. This knowledge is critical as perceptions of safety significantly impact cycling adoption, potentially hindering any city's strategy to increase cycling numbers if safety perceptions are not encompassed. 

In the future, we plan to expand the work here started. One possible way forward is to use identifiable image characteristics (e.g., using image segmentation or object detection) as predictors to rate environments' perception of safety scores directly. In turn, this approach would improve scaling even further, as environment characteristics and their impacts on the perception of safety could be computed without further comparisons from individuals. A second approach can be using this same information in covariate-based ranking methods. Third, an analysis can be made if different typologies of individuals (i.e., Geller's cycling profiles) have different perceptions of safety, which can help cycling promotion strategies to more accurately and effectively target some populations' needs. 

%%%%%%%%%%%%%%%%%%%%%%%%%%%%%%%%%%%%%%%%%%%%%%%%%%%%%%%%%%%%%%%%%%%%%%%%%%%%%%%%
%% 
%%%%%%%%%%%%%%%%%%%%%%%%%%%%%%%%%%%%%%%%%%%%%%%%%%%%%%%%%%%%%%%%%%%%%%%%%%%%%%%%
%\addtolength{\textheight}{-12cm}
\addtolength{\textheight}{-0cm}   % This command serves to balance the column lengths

%%%%%%%%%%%%%%%%%%%%%%%%%%%%%%%%%%%%%%%%%%%%%%%%%%%%%%%%%%%%%%%%%%%%%%%%%%%%%%%%
%% APPENDIX
% \textcolor{blue}{\blindtext[1]}

%%%%%%%%%%%%%%%%%%%%%%%%%%%%%%%%%%%%%%%%%%%%%%%%%%%%%%%%%%%%%%%%%%%%%%%%%%%%%%%%
%% ACKNOWLEDGMENT
%%%%%%%%%%%%%%%%%%%%%%%%%%%%%%%%%%%%%%%%%%%%%%%%%%%%%%%%%%%%%%%%%%%%%%%%%%%%%%%%
\section*{ACKNOWLEDGMENTS}
This work is part of the research activity partially funded by Fundação para a Ciência e Tecnologia (FCT) via grant [PD/BD/142948/2018] that was partially carried out at the Civil Engineering Research and Innovation for Sustainability (CERIS) funded by FCT [UIDB/04625/2020], the Associate Laboratory of Robotics and Engineering Systems (LARSyS) funded by FCT [UIDB/50009/2020], and the Department of Technology, Management, and Economics at the Technical University of Denmark (DTU).

%%%%%%%%%%%%%%%%%%%%%%%%%%%%%%%%%%%%%%%%%%%%%%%%%%%%%%%%%%%%%%%%%%%%%%%%%%%%%%%%
%% REFERENCES
%%%%%%%%%%%%%%%%%%%%%%%%%%%%%%%%%%%%%%%%%%%%%%%%%%%%%%%%%%%%%%%%%%%%%%%%%%%%%%%%

\bibliographystyle{IEEEtran}
\bibliography{IEEEabrv, references}

%%%%%%%%%%%%%%%%%%%%%%%%%%%%%%%%%%%%%%%%%%%%%%%%%%%%%%%%%%%%%%%%%%%%%%%%%%%%%%%%
%% END
%%%%%%%%%%%%%%%%%%%%%%%%%%%%%%%%%%%%%%%%%%%%%%%%%%%%%%%%%%%%%%%%%%%%%%%%%%%%%%%%
\end{document}